\documentclass[letterpaper,twocolumn,fleqn]{article} 

\usepackage{ist}
\usepackage{color, soul}
\usepackage{url}
\usepackage{comment}
\usepackage[sorting=none, firstinits, style=numeric-comp, doi=false, url=false, eprint=false, backend=biber, maxnames=1]{biblatex}
\usepackage{setspace}
\usepackage{listings}
\usepackage{listings}
\usepackage{xcolor}

\colorlet{punct}{red!60!black}
\definecolor{background}{HTML}{EEEEEE}
\definecolor{delim}{RGB}{20,105,176}
\colorlet{numb}{magenta!60!black}

\lstdefinelanguage{json}{
    basicstyle=\footnotesize\ttfamily,
    showstringspaces=false,
    breaklines=true,
    frame=lines,
    backgroundcolor=\color{background},
    literate=
     *{0}{{{\color{numb}0}}}{1}
      {1}{{{\color{numb}1}}}{1}
      {2}{{{\color{numb}2}}}{1}
      {3}{{{\color{numb}3}}}{1}
      {4}{{{\color{numb}4}}}{1}
      {5}{{{\color{numb}5}}}{1}
      {6}{{{\color{numb}6}}}{1}
      {7}{{{\color{numb}7}}}{1}
      {8}{{{\color{numb}8}}}{1}
      {9}{{{\color{numb}9}}}{1}
      {:}{{{\color{punct}{:}}}}{1}
      {,}{{{\color{punct}{,}}}}{1}
      {\{}{{{\color{delim}{\{}}}}{1}
      {\}}{{{\color{delim}{\}}}}}{1}
      {[}{{{\color{delim}{[}}}}{1}
      {]}{{{\color{delim}{]}}}}{1},
}

\title{AI-assisted Automated Workflow for Real-time X-ray Ptychography Data Analysis via Federated Resources}

\author{Anakha V Babu, Tekin Bicer, Saugat Kandel, Tao Zhou, Daniel J. Ching, Steven Henke, Sini\v{s}a Veseli, Ryan Chard, Antonino Miceli, Mathew Joseph Cherukara \\
Argonne National Laboratory, Lemont, Illinois, USA}

\date{} % date has an empty field.

\begin{document} 

\maketitle 

\thispagestyle{empty} % prevents the first page to be numbered

%%%%%%%%%%%%%%%%%%%%%%%%%%%%%%%%%%
% Abstract
%%%%%%%%%%%%%%%%%%%%%%%%%%%%%%%%%%

\begin{abstract}

%{\color{red} SK (proposed new abstract)

We present an end-to-end automated workflow that uses large-scale remote compute resources and an embedded GPU platform at the edge to enable AI/ML-accelerated real-time analysis of data collected for x-ray ptychography. Ptychography is a lensless method that is being used to image samples through a simultaneous numerical inversion of a large number of diffraction patterns from adjacent overlapping scan positions.  This acquisition method can enable nanoscale imaging with x-rays and electrons, but this often requires very large experimental datasets and commensurately high turnaround times, which can limit experimental capabilities such as real-time experimental steering and low-latency monitoring. In this work, we introduce a software system that can automate ptychography data analysis tasks. We accelerate the data analysis pipeline by using a  modified version of PtychoNN -- an ML-based approach to solve phase retrieval problem that shows two orders of magnitude speedup compared to traditional iterative methods.  Further, our system coordinates and overlaps different data analysis tasks to minimize synchronization overhead between different stages of the workflow. We evaluate our workflow system with real-world experimental workloads from the 26ID beamline at Advanced  Photon Source and ThetaGPU cluster at Argonne Leadership Computing Resources.

\end{abstract}

%%%%%%%%%%%%%%%%%%%%%%%%%%%%%%%%%%%%
% Overall Document Guidelines: Head
%%%%%%%%%%%%%%%%%%%%%%%%%%%%%%%%%%%%

%%%%%%%%%%%%%%%%%%%%%%%%%%%%%%%%%%
% Graphics and Equations
%%%%%%%%%%%%%%%%%%%%%%%%%%%%%%%%%%
\section{Introduction}

Neural networks (NN) are considered universal approximators and have been successful in solving problems in computer vision, natural language processing, autonomous control, and many others \cite{NN_ref}.  Machine learning (ML) based alternate approaches for inverse imaging problems have recently received significant attention and have shown considerable success for medical imaging, computational photography, computational microscopy, geophysical imaging, and other applications \cite{ongie_ijsaitt_2020}. 
For synchrotron radiation facilities specifically, ML-based reconstruction approaches have shown promise for x-ray imaging in the Bragg coherent diffraction imaging, high-energy diffraction microscopy, and ptychography modalities \cite{yao2022autophasenn, PtychoNN, BraggNN}. 
These ML-based approaches typically consist of two important components. First, the NN is trained using previously acquired (data, solution) pairs.  Second, the trained NN is deployed to invert new experimental data. 
This NN-based inversion is often a single-shot procedure that can be orders of magnitude faster than typical iterative inversion techniques, and this offers the possibility for real-time experimental feedback with high-resolution imaging data.

Ptychography is a particularly important  high-resolution coherent diffraction imaging (CDI) technique that has enabled nanoscale imaging in a range of x-ray and electron microscopy applications, including the imaging of integrated circuits~\cite{PSI-chip},  biological specimens like algae~\cite{Deng2015}, and strain imaging of nanowires~\cite{Hruszkewycz2017, Hill2018}, and even the atomistic (sub-angstrom) scale imaging of nanostructures~\cite{first_e_ptycho, second_e_ptycho}. 
In a typical x-ray ptychography experimental setup, a focused coherent x-ray beam is scanned across a sample and a large number of far-field intensity diffraction patterns are acquired at the detector. 
The nature of the CDI experiment is such that the phase information is lost and hence conventional iterative methods are used to retrieve the phase using the measured intensity. 
The iterative reconstruction techniques developed for ptychography are typically data and compute-intensive and require many (hundreds to thousands of) iterations over the measurement data~\cite{apsu}. 
With advances in accelerator technology, such as the upcoming upgrade of Argonne Photon Source (APS), the data rates of modern x-ray ptychography instruments are expected to increase by orders of magnitude, which, in turn, will require large-scale compute resources that can provide peta- to exa-FLOPS throughput to perform conventional ptychography analysis tasks~\cite{APS_comp_strategy}.
This raises prohibitive computational requirements for real-time data analysis using traditional  ptychography algorithms. 

The use of neural networks in real-time ptychographic reconstruction presents an attractive proposition, but it requires addressing some key logistical challenges. 
Considering the increased coherence and brilliance of next-generation light sources such as the APS-U, the NN training has to deal with huge amounts of data and will be performed at a remote high-performance computing (HPC) resource to achieve fast training by leveraging vast computational throughput. 
However, real-time feedback to the beamline is not possible with remote HPC for NN inference due to the latency involved in the data transfer~\cite{Anakha_thesis}. 
Thus it is essential to implement the inference at the edge or as close to the data generating source to enable real-time data analysis. 
With the availability of low-cost embedded GPUs and dedicated AI/ML hardware accelerators, the feed-forward computations in the NN are  feasible with high throughput~\cite{AI_Acc_survey}. 
The trained model can periodically be deployed to an edge device for low-latency operations, such as experimental steering and/or monitoring. 
Further, once the reconstructions of the trained model meet user quality metrics, valuable HPC resources can be released and used for other tasks. 
This motivates the need for developing a workflow system that will automatically monitor the data acquisition process and can efficiently use remote HPC resources for faster data processing, ML training, and edge inference. 
\section{Related Work}

% HPC algorithms for ptycho
Many reconstruction algorithms have been developed for ptychography over the years~\cite{nikitin2019photon, marchesini2016sharp,aslan2020distributed, shapiro2017ptychographic}. 
The parallelization of these algorithms and their efficient execution on high-end compute resources and accelerators, such as GPUs, have also gained attention~\cite{wang2022image, Nashed:14, shapiro2017ptychographic}. 
The 3D ptychography experiments further elevate the computational demands of the data analysis tasks, which extends the 2D ptychography with tomography or laminography techniques, increasing the size of collected datasets to several orders of magnitude~\cite{apsu, aps-raven}. 
Although most of the 3D reconstruction tasks are performed with single-pass direct methods (due to their computationally less demanding nature), high-quality 3D reconstructions still rely on iterative techniques.
Therefore parallel and distributed~\cite{bicer2017trace,bicer2015rapid}, memory-centric~\cite{hidayetouglu2019memxct}, high-performance 3D reconstruction algorithms~\cite{aslan2019joint} and techniques, which can provide quasi-real-time results~\cite{bicer2020tomostream}, have been developed~\cite{wang2017massively,hidayetoglu2020petascale}.
For extremely large-scale 3D datasets, parallelization of single-pass filtered-back projection methods is also studied~\cite{chen2019ifdk,chen2021scalable}.
However, the analysis of large-scale experimental datasets (for both direct and iterative methods) requires the usage of clusters and supercomputers, which are typically difficult to interact with. 
Several runtime and workflow systems have been developed in order to ease the usage of  these resources~\cite{vescovilinking,salim2019balsam,Basham:fv5032,donatelli2015camera}, their executions have been demonstrated for  synchrotron image analysis tasks~\cite{klein2019interactive,peterka2009configurable,bicer2017real}, including ptychography workflows~\cite{Tekin_workflow}.
Although these resources can provide massive computational throughput, they are still limited in terms of their feedback latency due to job scheduling, data movement, and resource allocation overheads; therefore, the usage of edge devices that are close to the data acquisition is desirable for low-latency tasks.  

% AI/ML-focused works
AI/ML techniques have been used to alleviate computational demands of ptychography data analysis and improve the output image quality, including for both 2D~\cite{aslan2021joint,kalinin2016big,venkatakrishnan2021algorithm} and 3D ptychography~\cite{liu2019deep,wu2020deep,liu2020tomogan}.
Surrogate models that replace the phase retrieval algorithms have recently shown promising results with very short inference times~\cite{PtychoNN, cherukara2018real,yao2022autophasenn}. 
These models have been deployed to edge computing devices, such as NVIDIA Jetson, that enable real-time feedback for scientists as well as other downstream tasks that require low-latency data analysis~\cite{edgePtychoNN_Anakha, BraggNN}.
The training of these AI/ML models, however, can be time-consuming and need high-end GPUs. Further, the variety of samples and features that need to be learned by these models can be massive, which necessitates continuous learning.

In this work, we complement existing work via incorporating AI/ML training with our workflow system~\cite{Tekin_workflow,edgePtychoNN_Anakha}. 
Specifically, we perform high-latency (continuous) training tasks at high-end compute resources, whereas the low-latency feedback operations are carried out at the beamline using an edge device (NVIDIA Jetson). 
The loosely-coupled execution between training and inference tasks enables us to provide close to real-time results while providing high-quality outputs with up-to-date AI/ML models.

\section{Objective}
This work focuses on the automation of ML-accelerated x-ray ptychography data analysis for 2D image
reconstruction and providing a live visualization of phase predictions made by the edge device located close to the data acquisition machine. 
The workflow system has several processing components, for instance, the data transfer between the local data acquisition system to the remote HPC and then initiating the reconstructions for training the ML model. Once the training is complete, the trained model is pushed to the edge device from the remote HPC. 
Our workflow system coordinates different processing components, for example, the communication between the data acquisition system and remote HPC, implementing remote function calls in the HPC resources using available resources, and performing model updates on the edge device without interrupting the data visualization. 
Under the hood, our system handles complex tasks, such as job scheduling, task synchronization, resource authentication and management.
\section{Methodology and Experimental Design}
\begin{figure*}
  \includegraphics[width=\textwidth]{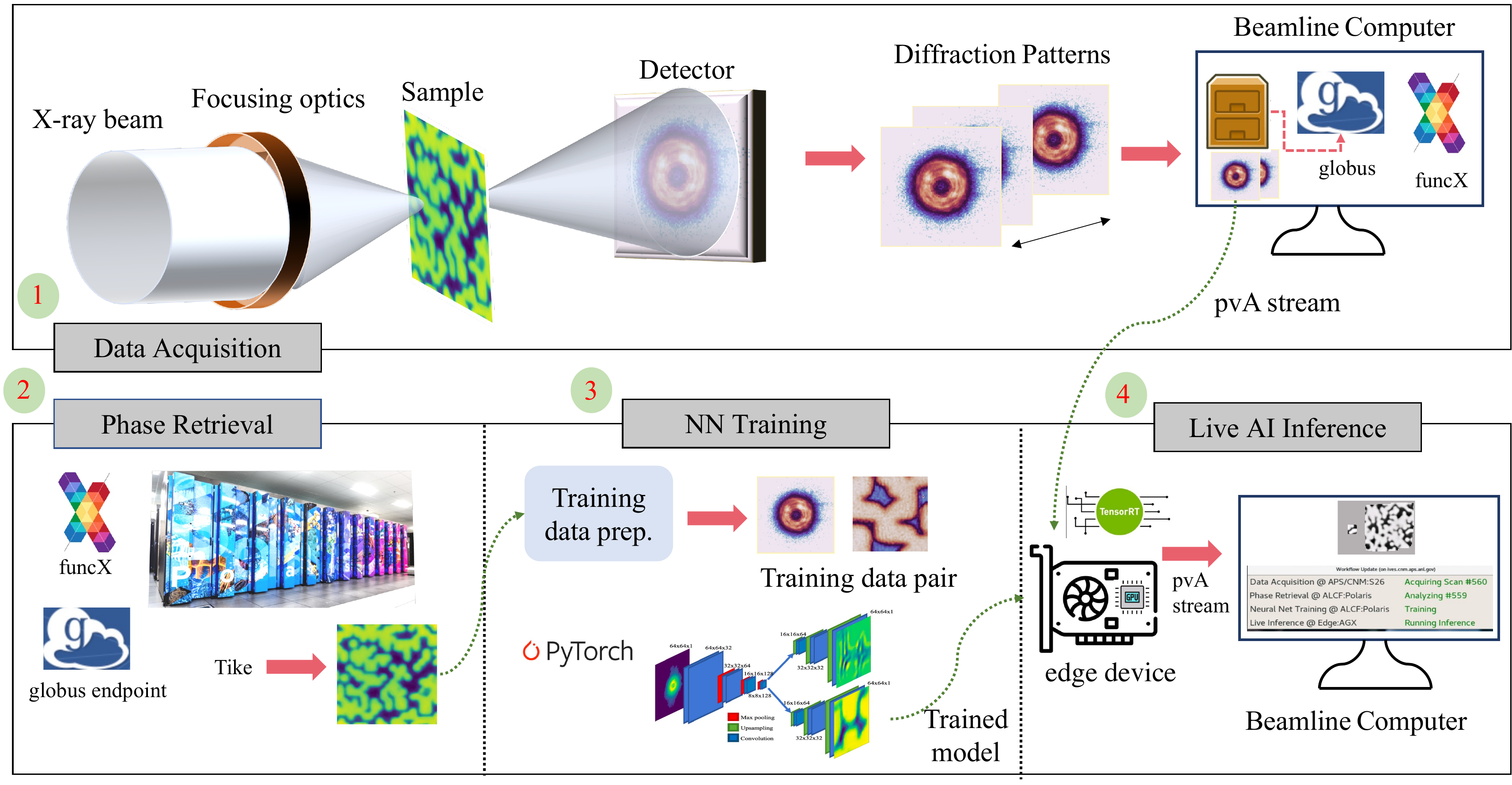}
  \vspace{-15px}
  \caption{An automated workflow for accelerating the x-ray ptychography data analysis using federated resources. Step (1) represents the data acquisition process during a ptychography experiment. 
  The acquired diffraction patterns are transferred to the HPC resources and fed into ptychographic reconstruction tasks as illustrated in step (2). 
  The reconstructed 2D real-space images are then used for training the PtychoNN model (step (3)). 
  The model training is a continuous process, i.e. the workflow system keeps reconstructing the real-space 2D images as the measurement data become available and used for training. 
  The trained up-to-date model is deployed to the beamline periodically for real-time inference as shown in step (4). 
  We use Globus tools and services to perform data transfer, remote function calls, authentication, and task and resource management.  }
  \label{Figure:workflow}
  \vspace{-15px}
\end{figure*}

The automated workflow for real-time ptychography data analysis can be visualized as shown in Figure \ref{Figure:workflow}. 
It has mainly four different stages: (i) data acquisition, (ii) phase retrieval, (iii) neural network training, and (iv) live AI/ML inference. We explain each of these stages in the following subsections.

\subsection{Data Acquisition}
The experiment was carried out on the hard x-ray nanoprobe beamline (26ID) at the APS. The photon energy was 10.4 keV. The sample was a 1.5 $\mu$m thick Tungsten film etched into random patterns. 
X-ray ptychographic images, the diffraction patterns, were collected using a medipix3 detector with 55 $\mu$m pixel size, sitting at 1.55 m downstream of the sample. 
A Fresnel zone plate with 160 $\mu$m diameter and 30 nm outermost zone width was used for the focusing. The optics were intentionally defocused by -150 $\mu$m to produce an overfocused beam with an FWHM of about 600 nm. Each ptychographic scan follows an optimized spiral path of 963 data points with a step size of 50 nm.

\subsection{Workflow System Components}
We implement the data analysis pipeline using Globus  tools; namely, Automate~\cite{chard2023globus}, funcX~\cite{chard2020funcx}, and Transfer~\cite{chard2014efficient} services.

\vspace{-10px}
\begin{lstlisting}[language=json, caption={Example of a Globus Automate flow definition.}, label={lst:flow-0}]
{"StartAt": "Init",      -> Beginning state
 "States": {             -> State definitions
   "Init": {             -> Init state
     "Type": "Action", 
     "ActionUrl": "https://.../transfer", ...
     "Parameters": {
       "src.$": "...", "dest.$": "...",
       "items": [...]
     }, ...
     "Next": "Analyze"
   }, ...                -> See Listing 3
\end{lstlisting}

% Globus Automate
The Globus Automation Services helps end users implement the steps (or stages) of the workflow using a JSON-based flow definition language. 
These stages consist of {\em action}s, which are collectively called {\em flow}s, including data transfer and remote function execution. 
These actions are typically triggered or executed via interacting with other services' application programming interfaces (APIs). 
An example of a flow definition is given in Listing~\ref{lst:flow-0}. 
The \texttt{StartAt} has the value of initial state for the workflow, which points to the \texttt{Init}. The following \texttt{States} entry consists of the definition of the workflow states with \texttt{Init} state defined at the beginning. 
The \texttt{Init} state performs a data transfer action, where the source, destination, and items (folders or files) are passed as parameters.
The \texttt{Next} key points to the next state in the workflow, which is \texttt{Analyze} (see Listing~\ref{lst:flow-funcx}).

Globus Automate uses the authentication infrastructure provided by Globus Auth, therefore the services that are integrated with Globus Auth can benefit from the single sign-on authentication layer~\cite{tuecke2016globus}.  
This capability significantly eases the interaction between many actions such as the management of data transfer, interaction with storage systems, and compute resources.
\vspace{-15px}
\begin{lstlisting}[language=json, caption={Function registration to funcX service and sample usage.}, label={lst:funcx}]
from funcx.sdk.client import FuncXClient
# Python function to be run
def ptycho_func(**data):  ...

main(): ...
 fxc = FuncXClient() -> Get funcX client to
 ...                    interact with service
 # Serialize the local python function and 
 # store it at funcX cloud service
 fxid = fxc.register_function(ptycho_func)
 fxep = getEndPointAddress()
 ...    
 # Execute ptycho_func at fxep endpoint
 fxc.run(params,           -> Param. to pass
         endpoint_id=fxep, -> Remote resource
         function_id=fxid) -> Func. to run
\end{lstlisting}

% FuncX
We use the funcX service 
% provided by Globus services 
to perform remote function execution. The remote execution of python functions using funcX involves two main steps: (1) registration of user-defined python functions to funcX cloud service and (2) execution of functions on the target compute resources. 
Step (2) requires interaction between funcX cloud service and endpoints as the user-defined functions need to be serialized, transferred, and scheduled for execution. 
These steps are shown in Listing~\ref{lst:funcx}. 
Specifically, \texttt{fxc.register\_function(...)} and \texttt{fxc.run(...)}. The funcX functions in Listing~\ref{lst:funcx} are used for registering and triggering the user-defined python function (\texttt{ptycho\_func(...)}). 
The funcX service resolves {\em which function} to run and {\em where to run} it with \texttt{functions\_id} and \texttt{endpoint\_id} parameters, respectively.
Registered funcX functions can be invoked within a Globus automation, as shown in Listing~\ref{lst:flow-funcx}.
The funcX endpoints manage the large-scale compute resources via the system's job scheduler.
\vspace{-10px}
\begin{lstlisting}[language=json, caption={Globus Automate funcX action implementation in flow definition.}, label={lst:flow-funcx}]
 ...                    -> See Listing 1
 "Analyze": {           -> Another state def.
   "Type": "Action", ...
   "ActionUrl": "https://automate.funcx.org",
   "Parameters": {
     "tasks": [{ "endpoint.$": "...",
       "function.$": "...",
       "payload.$": "..."
     }] },
   "Next": "Fin"        -> Next state
 },
 "Fin": { ...           -> Final state
   "End": True           
 }, #Fin } #States, }   -> End of state defs.
\end{lstlisting}

% Transfer
The workflow data transfer actions are performed by the Globus Transfer service. The transfer service, similar to the funcX service, handles the data movement operations using a cloud service that interacts with endpoints. 
The data transfer operations are fault tolerant and can handle extremely large data movements. 

Our workflow system automatically performs the transfer of the acquired ptychography data from beamline to compute cluster, initiates the reconstruction tasks, triggers the model training, and deploys the up-to-date model to the edge using Globus services (between facilities), workflow scripts (at the compute cluster), and pvAccess (at the beamline). Globus services seamlessly integrate with Globus Auth and provide secure access to large-scale high-end storage and compute resources.

\subsection{Phase Retrieval}
Step 1 in Fig.~\ref{Figure:workflow} illustrates a typical ptychography experimental setup that is used to acquire diffraction patterns. 
During an experiment, many of these diffraction patterns are collected which can result in TB-scale measurement data.
The ptychographic reconstruction process aims to recover the phase information and reconstruct 2D real-space images from  diffraction patterns. 
The data acquisition process can be modeled with $\psi^{i+1} = F(d, p, \psi^{i})$ formula. 
Here, $d$ corresponds to diffraction patterns (measured data on detector) with their corresponding scanning positions, $p$ is the probe information, and $\psi^{i}$ is the (initial) guess for the imaged sample~\cite{yu2022scalable}. 
The ptychographic reconstruction process tries to iteratively converge a set of values for $\psi^{i+1}$ that is consistent with the measurement data ($d$) while solving the phase retrieval problem.

We used Tike (version 0.22) \cite{tike} to implement the ptychographic reconstruction operations. Tike provides several iterative solvers, including conjugate gradient, difference map, least square, and regularized ptychographic iterative engine (rPIE), where we used rPIE solver during our workflow execution. 
Further, the solvers are parallelized and they can efficiently be executed on multiple GPUs and nodes~\cite{yu2021ptycho,yu2022scalable}. 
We initiate the reconstruction tasks via funcX service that calls an external Tike script with application parameters, such as the number of iterations and probes, initial estimates, and the frequency of writing reconstructed images. 

\subsection{Neural Network Training}
In this work, we use a slightly modified version of PtychoNN (version 2.0), a fully convolutional neural network discussed in \cite{PtychoNN} for achieving live AI inference at the edge.
We apply a continuous training scheme where the training data is increased as the experiment progresses (with the newly collected diffraction patterns included in the training dataset). 
Therefore, the model will not be trained from scratch for each experimental iteration, instead starts from the previously saved checkpoint of the model with the lowest validation loss. With 10\% of the training dataset reserved for validation, the neural network is trained using the PyTorch framework for 50 epochs adapting a cyclic learning rate policy. The learnable parameters of the model are updated using the adaptive moment estimation (ADAM) optimizer to minimize the absolute mean error (MAE) between the reference labels (reconstructions obtained from phase retrieval) and neural network outputs. 
The trained model with the lowest validation loss at the end of 50 epochs is deployed to the beamline (Jetson AGX device \cite{AGX}) by the workflow system.

\subsection{Live AI/ML Inference}
The live AI inference is handled by AGX Xavier \cite{AGX}, an embedded platform from NVIDIA as shown in step (4), Fig.~\ref{Figure:workflow}.
AGX Xavier (Jetson series)  has a compute capability of 32 TOPS and is designed to accelerate ML workloads for edge use cases. 
This device is physically located close to the data source or the synchrotron beamline and hosts an EPICS \cite{EPICS} pvAccess client to process the diffraction patterns that are streamed from the beamline computer using the pvAccess protocol \cite{PVA}. 
As discussed in the data acquisition stage, the diffraction patterns are concurrently streamed over the network from the beamline computer as a pvAccess stream in addition to saving the data into the file system. 
The incoming diffraction data at the edge device is processed by the pvAccess client to obtain a 2D image and then feed to the inference engine for predicting the phases. 
The inference engine uses the latest trained model from HPC and converts that into ONNX format, which is a hardware-independent intermediate representation. 
We used the TensorRT API \cite{TensorRT}, a library provided by NVIDIA to perform inference with reduced latency on NVIDIA's embedded GPU platforms.  
We used a batch size of 1 and the phase prediction obtained from the NN model is live-streamed to the beamline computer as a pvAccess stream. 
The live individual predictions and the overall stitched phase map of the sample are also provided at the user interface. 
A maximum frame rate of 100 fps is achieved with AGX Xavier as the beamline computer is limited by the network bandwidth of 1 Gbps for $512\times512$ pixels. 

We also extended the capability of live streaming at the edge when the detector is running at its peak acquisition rate of 2kHz. 

\section{Results and Discussion}
Our workflow system provides seamless integration between synchrotron radiation beamlines and compute clusters. Specifically, we tested our system on 26ID beamline at APS and performed reconstruction and model training at ALCF's ThetaGPU cluster. The trained models are periodically deployed to a NVIDIA Jetson edge device located at beamline.  

Our experiments show that the edge computational capability using Jetson AGX Xavier can achieve real-time feedback up to 200 fps, i.e., consuming and inferring 200 diffraction patterns per second,  while concurrently performing reconstruction and ML model training at the ALCF’s ThetaGPU cluster. 
We used Tike toolkit, which includes GPU-optimized reconstruction solvers, to perform efficient phase retrieval on ThetaGPU. 
The Tike solvers are called via funcX, which also handles task and resource management on ThetaGPU cluster.

We also previously demonstrated the workflow in a ptychography experiment using ALCF’s ThetaGPU cluster using up to 64 A100 GPUs (8 nodes) and 168 concurrent reconstruction workflows. 
This workflow can achieve a speedup of 29.6x compared to using a single A100 at the beamline with greater than 81\% compute scaling efficiency (as shown in \cite{Tekin_workflow}). The new workflow system extends our previous work with AI/ML capabilities, which further improves the turnaround latency.
If higher frame rates are desired at the edge, inference operations can be performed with multiple edge devices or other advanced NVIDIA GPUs. 
Our experiments show that the inference rate can reach up to 8000 fps using 2 RTX A6000 GPUs.

Our workflow system provides basic building blocks for inter-facility integration via Globus tools and services. This eases the extension of workflow with additional steps (as new states in Globus Automate). Similarly, other data analysis workflows can easily be implemented with these building blocks~\cite{vescovilinking}.

\section{Acknowledgements}
%Argonne National Laboratory’s work was supported by the U.S. Department of Energy, Office of Science, Office of Basic Energy Sciences and Advanced Scientific Computing Research, under contract DE-AC02-06CH11357. 
Work performed at the Advanced Photon Source, Argonne Leadership Computing
Facility and the Center for Nanoscale Materials, all three U.S. Department of Energy Office of Science User Facilities, was supported by the U.S. DOE, Office of Basic Energy Sciences, under Contract No. DE-AC02-06CH11357.
We also acknowledge support from ANL's Laboratory Directed Research and Development funding 2023-0104.

\begingroup%\footnotesize
\renewcommand*{\bibfont}{\footnotesize}
\setstretch{0.95}
\setlength{\bibitemsep}{0.5\itemsep}
\printbibliography
\endgroup
%\bibliographystyle{unsrt}
%\bibliography{references,ptycho}

\begin{comment}
\comment{

}
\end{comment}
%%%%%%%%%%%%%%%%%%%%%%%%%%%%%%%%%%
% Biography
%%%%%%%%%%%%%%%%%%%%%%%%%%%%%%%%%%

\begin{biography}

Anakha V Babu  received the B.Tech degree in Electronics and Communication Engineering from University of Kerala, India, in 2010 and M.tech degree in Microelectronics \& VLSI design from the Indian Institute of Technology Bombay, Mumbai, India, in 2016. She completed her Ph.D. degree in Electrical Engineering at the New Jersey Institute of Technology, New Jersey, USA, in 2020. She worked as a postdoctoral appointee at the X-ray Science Division, Argonne National Laboratory (ANL), where she focused on implementing and automating real-time ptychography data analysis at the edge. She is currently working as a Hardware Electronics Engineer at the Beamline Controls and Data Acquisition (BCDA) of the APS, ANL. 

Tekin Bicer is a computer scientist in Data Science and Learning division at ANL. He has expertise in HPC, large-scale parallel and distributed systems, and AI/ML methods, with a special focus on large-scale x-ray image analysis problems. He received his Ph.D. from Computer Science and Engineering Department at Ohio State University in 2014, where he studied large-scale computing systems that address data management and analysis problems on high-end clusters and cloud compute resources.

Saugat Kandel received his B.A in Physics from Amherst College in 2008, and his Ph.D. in Applied Physics from Northwestern University in 2021. He has since been working on AI and optimization approaches for ptychography and autonomous experimentation methods in the Computational X-ray Sciences group at the APS in ANL.

Tao Zhou is an assistant scientist at the Center for Nanoscale Materials of ANL as well as a beamline scientist on the hard x-ray nanoprobe beamline at the APS. He received his PhD in Physics from the University of Grenoble Alpes in 2016. His main research focus is the development of novel x-ray diffraction imaging techniques assisted by advanced computation and machine learning methods.

Daniel J. Ching received his Ph.D. for Materials Science from Oregon State University in 2018. He has since been working as a computational imaging researcher for tomography and ptychography at ANL.

Steven Henke received his Ph.D. in Computational Science from Florida State University in 2013 and his B.S. degree in Physics from University of Wisconsin. He is a Data Scientist at APS Scientific Software Engineering since 2021.

Sini\v{s}a Veseli received his B.S. degree from the University of Zagreb, Croatia, in 1992. He received his Ph.D. in Physics at  the University of Wisconsin-Madison in 1996.
He is currently a member of APS Scientific Software Engineering and Data Management Group, and working on the controls Data Acquisition software for the APS Upgrade, as well as on the APS Data Management project.

Ryan Chard received his Ph.D. in computer science from Victoria University of Wellington, New Zealand. After an appointment as a Maria Goeppert Mayer Fellow at Argonne National Laboratory, he joined the Globus team to work on the development of cyberinfrastructure to enable scientific research.

Antonino Miceli is the Group Leader of the Detectors group in the X-ray Science Division of the APS at ANL with joint appointments at University of Chicago and Northwestern University.  He conducts R\&D in x-ray detectors to enable scientific advances, and operates the APS Detector Pool, which provides technical services and support for APS beamlines and users. Dr. Miceli received his Ph.D. in physics from the University of Washington. He joined ANL in 2005.

Mathew Cherukara Joseph  received his Ph.D. in computational materials science and engineering from Purdue University, and a bachelors in materials engineering from the Indian Institute of Technology Madras. He is currently the Group Leader of the Computational X-ray Science group at the APS. His research leverages AI for autonomous experimentation, materials characterization (metrology) beyond hardware limits and accelerated materials modeling.

\end{biography}

\section*{License}
\noindent
The submitted manuscript has been created by UChicago Argonne, LLC, Operator of Argonne National Laboratory (``Argonne"). Argonne, a U.S. Department of Energy Office of Science laboratory, is operated under Contract No. DE-AC02-06CH11357. The U.S. Government retains for itself, and others acting on its behalf, a paid-up nonexclusive, irrevocable worldwide license in said article to reproduce, prepare derivative works, distribute copies to the public, and perform publicly and display publicly, by or on behalf of the Government. The Department of Energy will provide public access to these results of federally sponsored research in accordance with the DOE Public Access Plan. http://energy.gov/downloads/doe-public-access-plan.

\end{document}